\documentclass[runningheads,a4paper]{llncs}

\usepackage{amssymb}
\usepackage{graphicx}
\usepackage{subfigure}

\usepackage{url}
\newcommand{\keywords}[1]{\par\addvspace\baselineskip
\noindent\keywordname\enspace\ignorespaces#1}

\begin{document}

\mainmatter  % start of an individual contribution

\title{Graph-Based Approaches to Clustering Network-Constrained Trajectory Data}

\titlerunning{Clustering Network-Constrained Trajectory Data}

\author{Mohamed K. El Mahrsi\inst{1} \and Fabrice Rossi\inst{2}}

\institute{T\'el\'ecom ParisTech, D\'epartement INFRES\\
46, rue Barrault 75634 Paris CEDEX 13, France\\
\email{khalil.mahrsi@telecom-paristech.fr}
\and
\'Equipe SAMM EA 4543, Universit\'e Paris I Panth\'eon-Sorbonne\\
90, rue de Tolbiac 75634 Paris CEDEX 13, France\\
\email{fabrice.rossi@univ-paris1.fr}
}

\maketitle

%\pagenumbering{gobble}

\begin{abstract}
Even though clustering trajectory data attracted considerable attention in the last few years, most of prior work assumed that moving objects can move freely in an euclidean space and did not consider the eventual presence of an underlying road network and its influence on evaluating the similarity between trajectories. In this paper, we present two approaches to clustering network-constrained trajectory data. The first approach discovers clusters of trajectories that traveled along the same parts of the road network. The second approach is segment-oriented and aims to group together road segments based on trajectories that they have in common. Both approaches use a graph model to depict the interactions between observations w.r.t. their similarity and cluster this similarity graph using a community detection algorithm. We also present experimental results obtained on synthetic data to showcase our propositions.

\end{abstract}

\section{Introduction}

Recent progress in telecommunications and geo-positioning contributed to the democratization of location-aware devices (GPS, smartphones, PDA, etc.) capable of retrieving, storing and sharing their position. Thanks to these devices, it becomes feasible to construct dedicated systems to store the trajectories of various types of moving objects (vehicles, pedestrians, etc.). The availability of such data has shed the light on new challenges and motivated work on management, analysis and data mining of Moving Object Databases (\textit{MOD}) \cite{Giannotti_2008}.

One of the domains where mining trajectory data can be beneficially applied is road traffic analysis. In fact, collecting and analyzing the GPS logs of vehicles traveling along the road network in order to deduce the state of the network and understand the flow dynamics can be a more efficient and affordable alternative to the use of dedicated sensors (which are expensive to deploy and maintain).

This paper tackles the problem of clustering trajectories of vehicles moving along a road network. More precisely, we study two separate clustering problems: i. the trajectory-based clustering problem, in which we are interested in discovering groups of trajectories that visited the same parts of the road network; and ii. the segment-based clustering problem, which aims to retrieve clusters of road segments that are co-traveled frequently. Studying those clustering problems can be useful in many contexts, such as:
\begin{itemize}
	\item Management and planning of road network infrastructure: trajectory clustering gives insight into the way the road network is used and grouping together segments that are co-traveled on regular basis can be very helpful for predicting propagation of congestion situations. This knowledge can be used to guide future planning decisions (construction of new roads and lanes, etc.);
	\item Carpooling: clusters of trajectories represent opportunities for carpooling which, if seized, can help reduce traveling costs and have a positive environmental impact.
\end{itemize}

The majority of existing literature consider the case of objects moving freely on the euclidean space, neglecting, in the case of vehicular trajectories, the movement constraints imposed by the underlying road network. Moreover, existing approaches are often very sensitive to their parameter values which should be fine-tuned depending on the dataset at hand to produce relevant clustering results. Our contribution addresses these two concerns and can be summarized in the following points:
\begin{itemize}
	\item We define a similarity measure that compares trajectories based on their proximity on the road network;
	\item We propose a graph representation to model interactions between trajectories w.r.t. their similarity.
	\item We use modularity-based community detection to cluster the similarity graphs and discover hierarchies of nested trajectory clusters suitable for exploration at various levels of detail;
	\item We proceed in analogy and define a clustering approach that regroups road segments based on common trajectories that travel them;
	\item We present experimental results to showcase our segment clustering approach on a synthetic dataset.
\end{itemize}

The rest of this paper is organized as follows. We present our data model and our formulation of the studied clustering problems in Section \ref{sec:DataModel}. Section \ref{sec:TrajectoryClustering} describes our trajectory clustering approach. Our road segment clustering approach is presented in Section \ref{sec:SegmentClustering}. Experimental results are shown in Section \ref{sec:Experiments}. Related work is discussed in Section \ref{sec:RelatedWork}. Finally, Section \ref{sec:Conclusion} concludes the paper.

\section{Data Representation and Problem Statement}
\label{sec:DataModel}

A road network is, the most commonly, represented as a directed graph $G = (V, E)$ \cite{Kharrat_2008,Lou_2009,Roh_2010}. The set of nodes, or vertices, $V$ represents intersections and terminal points of roads whereas the set of directed edges $E$ represents the road segments interconnecting those nodes. A directed edge $e = (v_i, v_j)$ indicates that a road segment links the two nodes $v_i$ and $v_j$ and that it can be traveled from $v_i$ in the direction of $v_j$ but not the other way around (unless another edge states otherwise).

A constrained trajectory $T$ that travels along this road network can be modeled as a sequence of connected segments:
\begin{equation}
T = \left< id, \left\{e_1, e_2,  ... , e_l\right\} \right>
\end{equation}

Where $id$ is the identifier of the trajectory, $l$ its length (i.e. number of segments) and $\forall 1 \leq i < l, e_i$ and $e_{i+1}$ are connected segments belonging to $E$. In a real-case scenario, trajectories are collected as GPS logs (sequences of latitude and longitude points) on which a map matching technique (e.g. \cite{Brakatsoulas_2005,Lou_2009}) is applied in order to produce the sequence of traveled segments. The map matching step is out of the scope of this paper.

Given a set of trajectories $\mathcal{T} = \left\{ T_1, T_2, ... , T_n \right\}$ that traveled along a road network $G = (V, E)$, we define the following two clustering problems that we will try to solve afterwards:
\begin{itemize}
	\item The trajectory-based clustering problem consists in partitioning $\mathcal{T}$ into a set of clusters $\mathcal{C} = \{C_1, ..., C_K\}$ of trajectories that exhibited similar behavior. Resemblance between trajectories of the same cluster $C_i$ should be as high as possible and trajectories across two different clusters $C_i$ and $C_j$ should be as different as possible.
	\item The segment-based clustering problem, on the other hand, aims to partition the set of road segments $E$ into clusters of segments that recurrently co-appear in  trajectories. It is desirable that: i. if a segment $s$ in a given cluster $C_i$ appears in a subset of trajectories in $\mathcal{T}$ then other segments in $C_i$ are likely to appear in that same subset; and ii. segments belonging to different clusters are unlikely to appear together very often.
\end{itemize}

\section{Trajectory-Based Clustering}
\label{sec:TrajectoryClustering}

In this section, we present our solution to the trajectory-based clustering problem. Our clustering framework can be divided into three steps. First, we introduce a measure to assess the similarity between trajectories based on their proximity on the road network (Section \ref{sec:similarity}). Then, we use this measure to build a graph depicting the similarity between trajectories (Section \ref{sec:graph}). This similarity graph is partitioned using a community-detection algorithm (Section \ref{sec:clustering}) in order to discover meaningful trajectory clusters.

\subsection{Similarity Between Trajectories}
\label{sec:similarity}

We adopt a bag-of-segments model where each trajectory is considered as an unordered collection of segments. When comparing two trajectories, the presence of each segment is checked individually without accounting for the segment's order in the trajectory or the presence of other segments. This simplification is justified by two observations: i. even if the order is unaccounted for explicitly, the underlying network model is a directed graph. Consequently, the direction of travel is implicitly respected since the visited edges are not the same for each direction; and ii. in a context of traffic analysis, congestion situations occur first in singular isolated segments and spread afterwards to adjacent segments, considering individual segments as the basis for comparison is the most natural and intuitive choice.

Intuitively, all segments do not have the same discriminative power. Segments that are frequently traveled by the majority of trajectories are not very relevant to cluster formations. On the contrary, segments that are traveled by a small portion of trajectories play a key role in the formation of the cluster containing those trajectories. To account for this observation, we devise a segment weighting strategy by adapting the TF-IDF (Term Frequency - Inverse Document Frequency) weighting to the case of trajectory data.

We define the spatial segment frequency (ssf) to measure the importance of an edge $e$ in a trajectory $T$:
\begin{equation}
	\mbox{ssf}_{e,T} = \frac{n_{e, T} \cdot \mbox{length}(e)}{\sum_{e' \in T} n_{e', T} \cdot \mbox{length}(e')}
\end{equation}

$n_{e,T}$ being the number of occurrences of $e$ in $T$ ($n_{e,T} = 1$ most of the time as trajectories rarely visit a segment more than once) and $length(e)$ its spatial length.

The inverse trajectory frequency (itf) measures the frequency of the segment $e$ in the whole set of trajectories $\mathcal{T}$:
\begin{equation}
	\mbox{itf}_e = \log\frac{| \mathcal{T} |}{| \{T_i : e \in T_i \} |}
\end{equation}

$| \mathcal{T} |$ is the total number of trajectories in the set $\mathcal{T}$ and $| \{T_i : e \in T_i \} |$ the number of trajectories containing the segment $e$. While inspecting trajectory $T$, the weight attributed to the road segment $e$ is the combination of both its ssf in the trajectory and its itf:
\begin{equation}
	\omega_{e,T} = \mbox{ssf}_{e,T} \cdot \mbox{itf}_e
\end{equation}

Finally, we compare two trajectories $T_i$ and $T_j$ by calculating their cosine similarity:
\begin{equation}
\label{eq:cos}
\mbox{Similarity}(T_i,T_j) = \frac{\sum_{e \in E} \omega_{e,T_i} \cdot \omega_{e,T_j}}{\sqrt{\sum_{e \in E} \omega_{e,T_i}^2} \cdot \sqrt{\sum_{e \in E} \omega_{e,T_j}^2}}
\end{equation}

\subsection{Trajectory Similarity Graph}
\label{sec:graph}

\begin{figure}[htdp]
	\begin{center}
		\includegraphics[width = .5\textwidth]{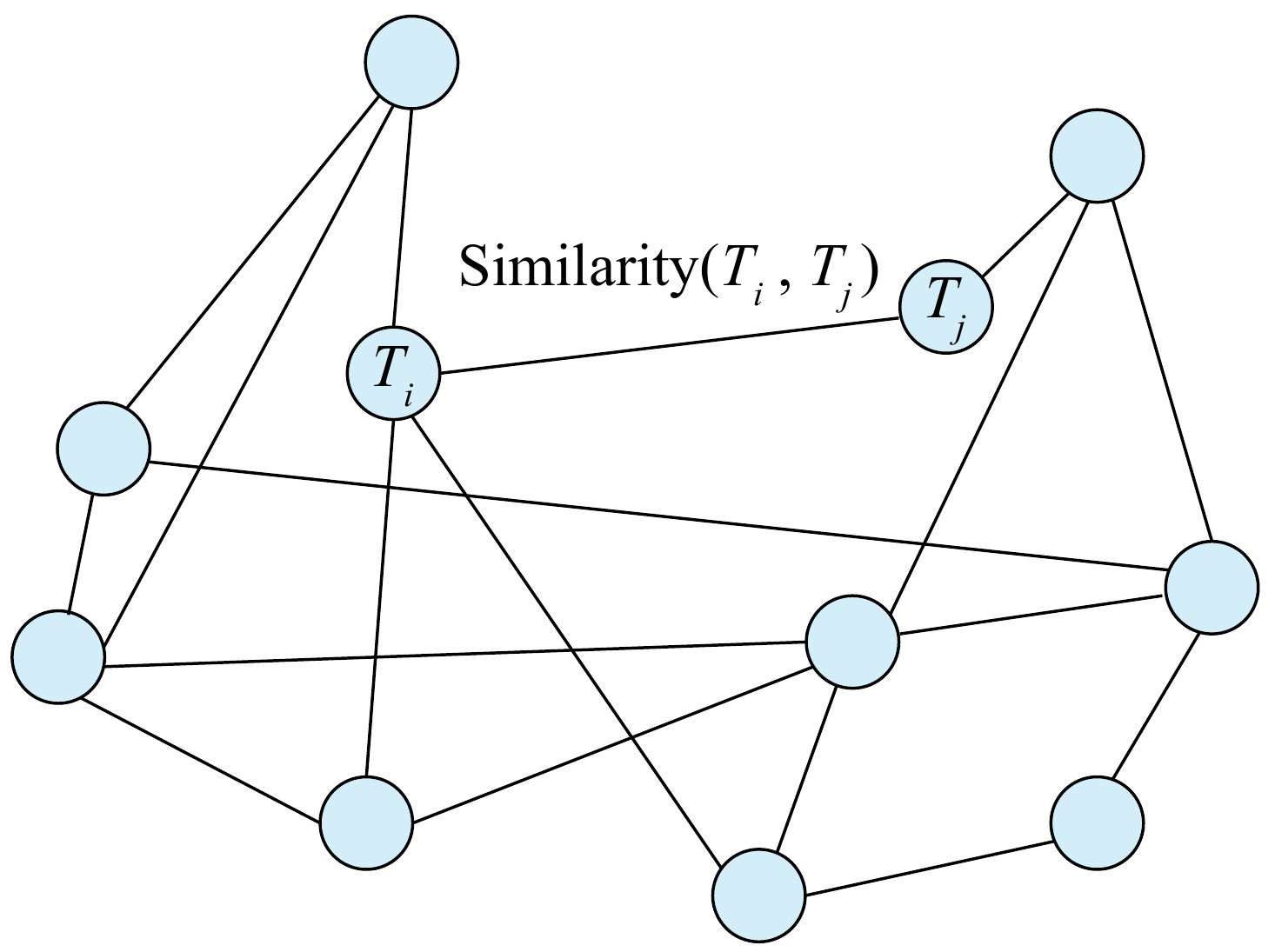}
	\caption{Example of a trajectory similarity graph: nodes represent trajectories while edges represent the intensity of the similarity between the trajectories.}
	\label{fig:simgraph}
	\end{center}
\end{figure}

We model the similarity relationships between trajectories as an undirected, weighted graph $G_{\mbox{S}} = (\mathcal{T}, E', W)$. Each trajectory in the set $\mathcal{T}$ is modeled as a node in $G_{\mbox{S}}$. An edge $e \in E'$ between a pair of trajectories $T_i$ and $T_j$ exists if and only if $\mbox{Similarity}(T_i, T_j) > 0$ (i.e. if they share at least one common road segment). In this case, the similarity is assigned as a weight $\omega_e \in W$ to the edge $e$. The concept of similarity graph is depicted in Fig.\ref{fig:simgraph}.

In addition to being a natural representation of the interactions between trajectories, the similarity graph puts extra emphasis on the fact that two trajectories that do not share common road segments are totally independent and are not to be "immediately" grouped together in a same cluster. This is reflected in the absence of a similarity edge linking the two trajectories and is fundamentally different from existing approaches that use classic distances and similarities and which can result in regrouping such trajectories together (especially those based on an unconstrained model, which can confuse dissimilar trajectories traveling along two separate but parallel and close roads).

Let $n = |\mathcal{T}|$ be the number of trajectories and $m = |E|$ the number of segments in the road network's graph $G$. The computational complexity for constructing the similarity graph is $O(mn^2)$ since it requires $\frac{n (n-1)}{2}$ similarity calculations each costing $O(m)$ at the very most.

\subsection{Clustering the Similarity Graph}
\label{sec:clustering}
Clustering is generally conducted on massive volumes of data. Therefore, the similarity graph described in the previous section tends to have a large number of nodes and edges (especially since only one common road segment is sufficient to create a similarity edge between two trajectories). Modularity-based community-detection algorithms are a popular and widely adopted choice to cluster such graphs \cite{Fortunato_2010}. Given a graph $G = (V, E, W)$, with vertices $V = \{v_1, v_2, ... , v_n\}$, weighted edges $E$ such as $\omega_{ij} \geq 0$ and $\omega_{ij} = \omega_{ji}$, and given a partition of the vertices into $K$ clusters (or communities) $C_1, ... , C_K$, the modularity of the partition is expressed according to the formula:

\begin{equation}
\mathcal{Q} = \frac{1}{2m}\sum_{k = 1}^K \sum_{i, j \in C_k} \left( \omega_{ij} - \frac{d_i d_j}{2m} \right)
\end{equation}
$d_i = \sum_{j \neq i} \omega_{ij}$ and $m = \frac{1}{2}\sum_i d_i$. The modularity measures the quality of the clustering by inspecting the arrangement of the edges within the communities of vertices. A high modularity is an indicator that the edges within the communities outnumber (or have higher weights than) those in a similar randomly generated graph (that does not present a community structure). Communities discovered using modularity optimization have a structure that is similar to the structure of cliques. In our case, this means that trajectories grouped together are heavily connected (which is the intended result) and share, two at a time, a considerable number of significant road segments. Moreover, the modularity accounts for the inspected nodes' degrees and can therefore avoid problems that might occur when dealing with lengthy trajectories (that are connected to a great number of other trajectories).

To cluster our similarity graph, we use an implementation of the algorithm described in \cite{Noack_2009} (the full details of the used implementation as well as the pseudo-codes of its different steps can be found in \cite{Rossi_2011}). First, the algorithm finds, for the similarity graph $G_{\mbox{S}}$, the partition $\mathcal{C} = \{C_1, C_2, ... , C_K\}$ that has the highest modularity. The significance of this partition is then evaluated by comparing its modularity to the modularity of optimal partitions obtained on similar randomly generated graphs. If the partition is valid (i.e. presents indeed a community structure), then the clustering algorithm is applied recursively on each of its clusters: for each cluster $C_i \in \mathcal{C}$, the sub-graph containing only nodes in $C_i$ and their similarity edges is isolated and clustered separately. The recursion stops once no further significant partitions can be found. The result of this clustering step is a complete hierarchy of nested trajectory clusters that can be explored at various levels of detail.

Since the similarity graph contains $n$ nodes and, at most, $\frac{n (n-1)}{2}$ edges ($n$ being the number of trajectories in $\mathcal{T}$), the theoretical (maximal) complexity of the community detection algorithm used in our clustering phase is $O(n^3)$ as reported in \cite{Fortunato_2010} (however, this complexity is rarely observed in practice where the complexity is somewhere near $O(n^2)$).

\section{Segment-Based Clustering}
\label{sec:SegmentClustering}

In this section, we are interested in discovering groups of segments that are very often traveled together. To solve our segment-based clustering problem, we proceed in analogy to the approach presented in the previous section. We define a similarity measure between segments based on comparison of the trajectories that travel them. This similarity is used to construct a similarity graph that is partitioned in order to discover the clusters of segments.

\subsection{Similarity Between Road Segments}

We consider that two road segments represented by the two directed edges $e_i$ and $e_j$ ($e_i, e_j \in E$) are similar if they frequently co-appear in the same trajectories (i.e. there exists a non-empty subset of trajectories in $\mathcal{T}$: $\{T \in \mathcal{T} : e_i \in T \wedge e_j \in T\}$). The larger the number of concomitant appearances of both segments is, the more they are considered similar. We can assimilate this concept of segment similarity to considering each segment as the bag-of-trajectories that traveled it. Comparing two road segments is, therefore, equivalent to comparing the collections of trajectories that visited each one of them.

The claim that we advanced in Section \ref{sec:similarity} can be extended to the case of comparing road segments. A lengthy trajectory that visits a considerable number of road segments is not very informative when judging the similarity between two segments in particular and vice versa. Therefore, we assign weights to trajectories depending on their contribution in characterizing road segments. The weight of a trajectory $T$, while inspecting a road segment $e$, is:

\begin{equation}
	\omega_{T, e} = \frac{n_{e,T}}{\sum_{T' \in \mathcal{T}} n_{e,T'}} \cdot \log\frac{|E|}{|e \in E: e \in T|}
\end{equation}

The first term in this weight calculates the contribution of $T$ to the segment $e$ by calculating the ratio between the number of appearances of $e$ in $T$ and the total number of appearances of $e$ in the whole dataset of trajectories $\mathcal{T}$. The second term evaluates the general importance of the trajectory and drops as the trajectory visits more segments and vice versa. Here again, we compare segments using their cosine similarity (by analogy to the formula in eq.(\ref{eq:cos})).

\subsection{Constructing and Clustering the Segment Similarity Graph}

We define the segment similarity graph in the same fashion that we defined the trajectory similarity graph in Section \ref{sec:graph}. The segment similarity graph $G_S(E, E', W)$ is an undirected, weighted graph where each node represents a road segment $e \in E$. A similarity edge $e' \in E'$ links two edges $e_i$ and $e_j$ if $Similarity(e_i, e_j) > 0$ (i.e. if $\{ T \in \mathcal{T}:  e_i \in T \wedge e_j \in T\}$ contains at least one trajectory) in which case their similarity is assigned as a weight to $e'$.

We refer to this approach of constructing the similarity graph as the "loose" approach. We can also devise a "strict" approach where a similarity edge exists between the two edges $e_i$ and $e_j$ not only if $Similarity(e_i, e_j) > 0$ but also if $e_i$ and $e_j$ are connected (i.e. the end node of one of the segments is the start node of the other). We apply the exact same community-detection algorithm mentioned in Section \ref{sec:clustering} to the segment similarity graph in order to discover the clusters of similar road segments. The theoretical complexities of the graph construction and the clustering steps can be deduced by analogy from those mentioned in Section \ref{sec:TrajectoryClustering}.

\section{Experimental Results}
\label{sec:Experiments}

In this section, we present preliminary results of the segment-based clustering approach on a synthetic dataset simulated using the Brinkhoff generator \cite{Brinkhoff_2002} (the interested reader is referred to \cite{Mahrsi_2012a} where we briefly reported some results of our trajectory-based approach). The dataset was generated using the Oldenburg map, which contains 6105 nodes and 14070 directed edges. It contains 1000 trajectories that traveled along a total of 7890 unique segments.

\begin{figure}[h]
\centering
\subfigure[]{
		\includegraphics[scale=0.35]{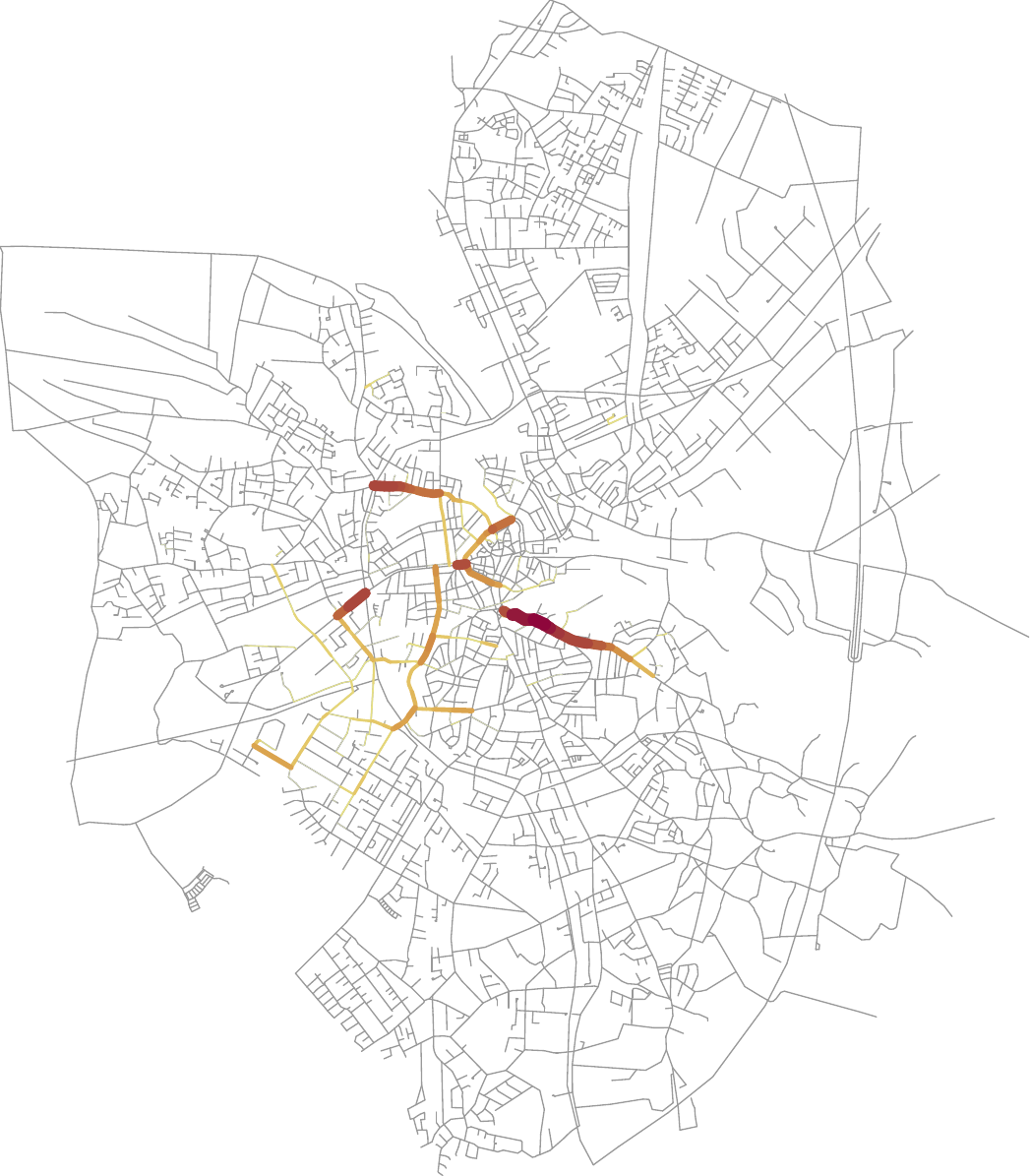}
}
\subfigure[]{
		\includegraphics[scale=0.35]{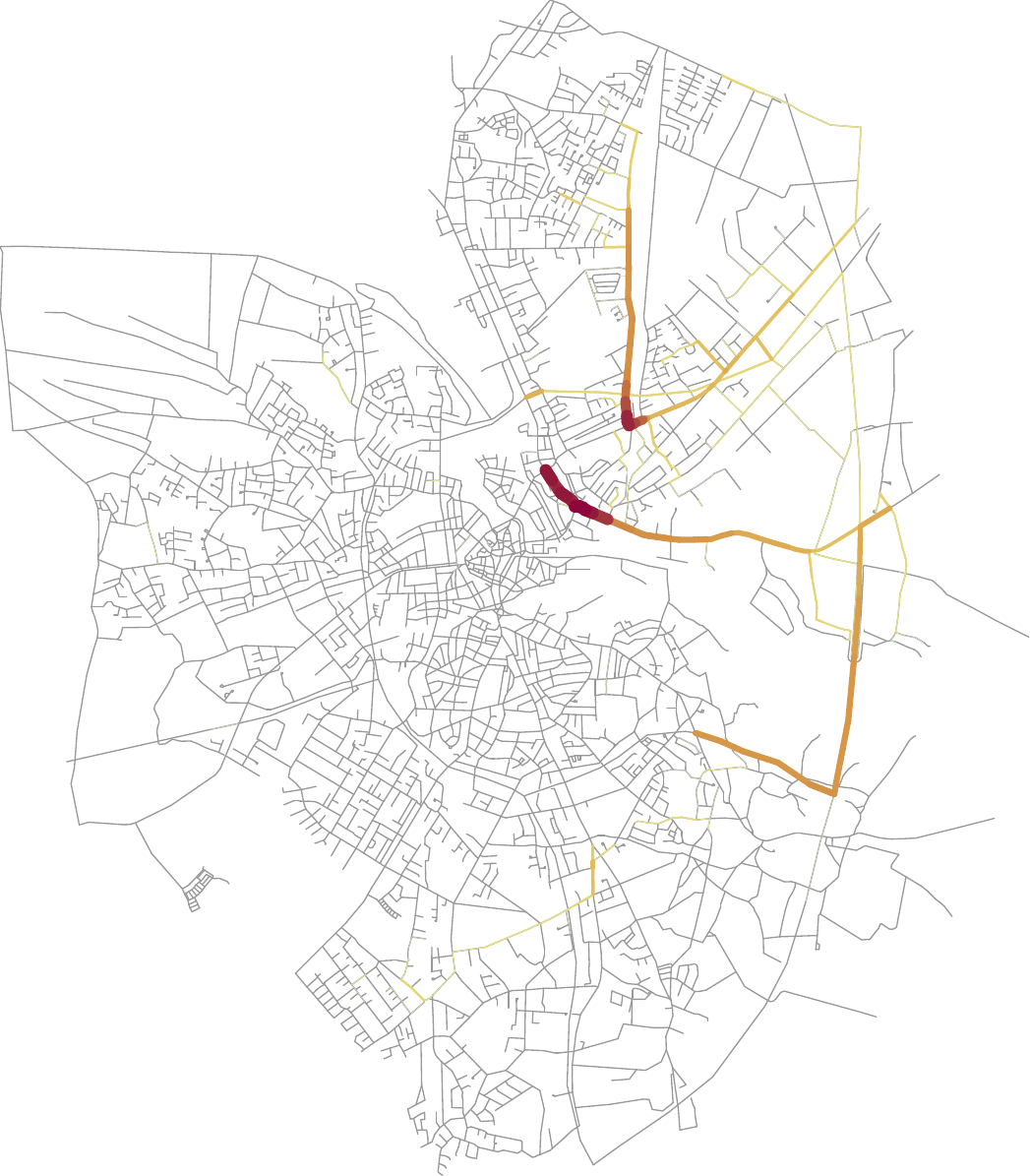}
}
\subfigure[]{
		\includegraphics[scale=0.35]{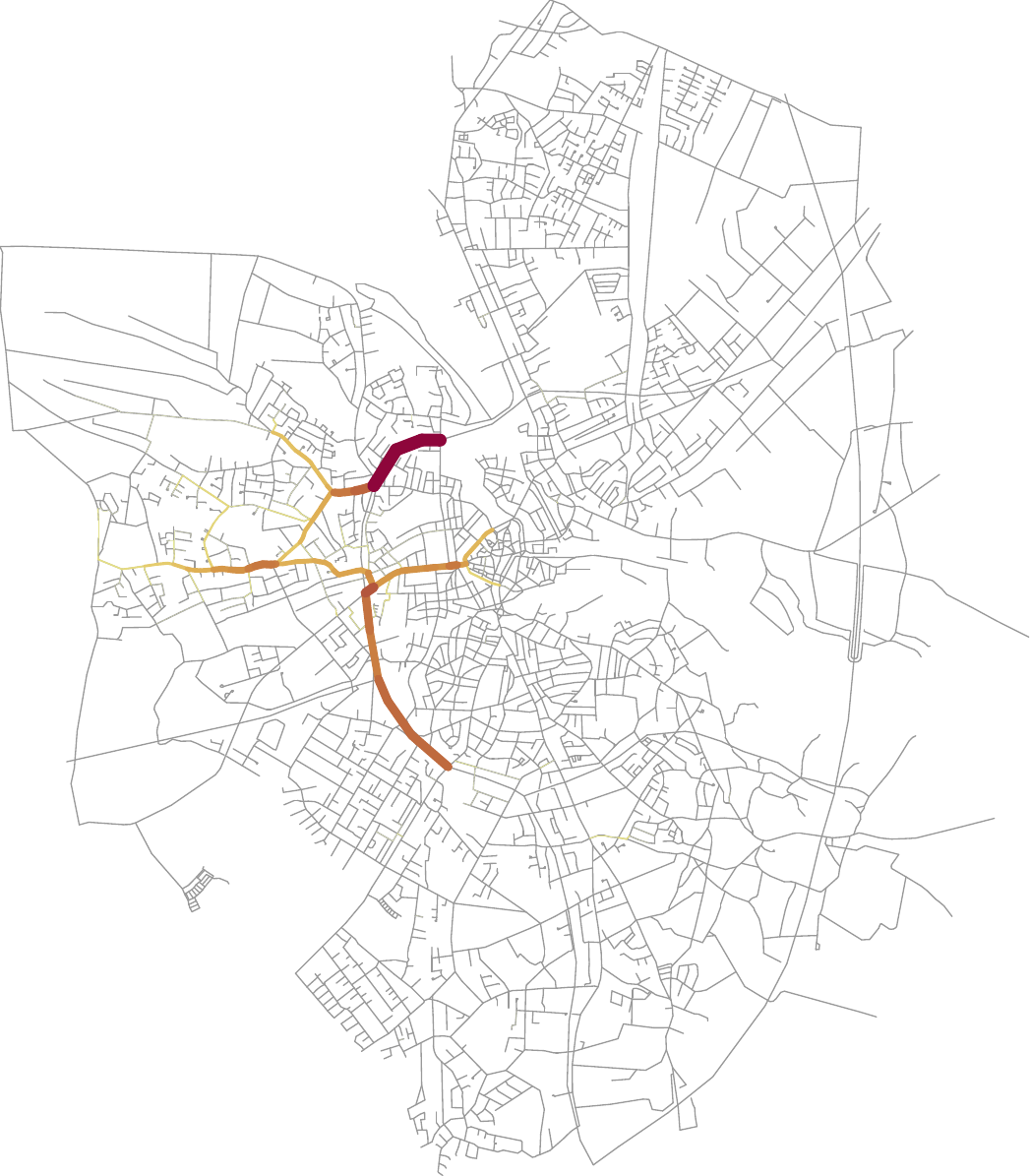}
}
\subfigure[]{
		\includegraphics[scale=0.35]{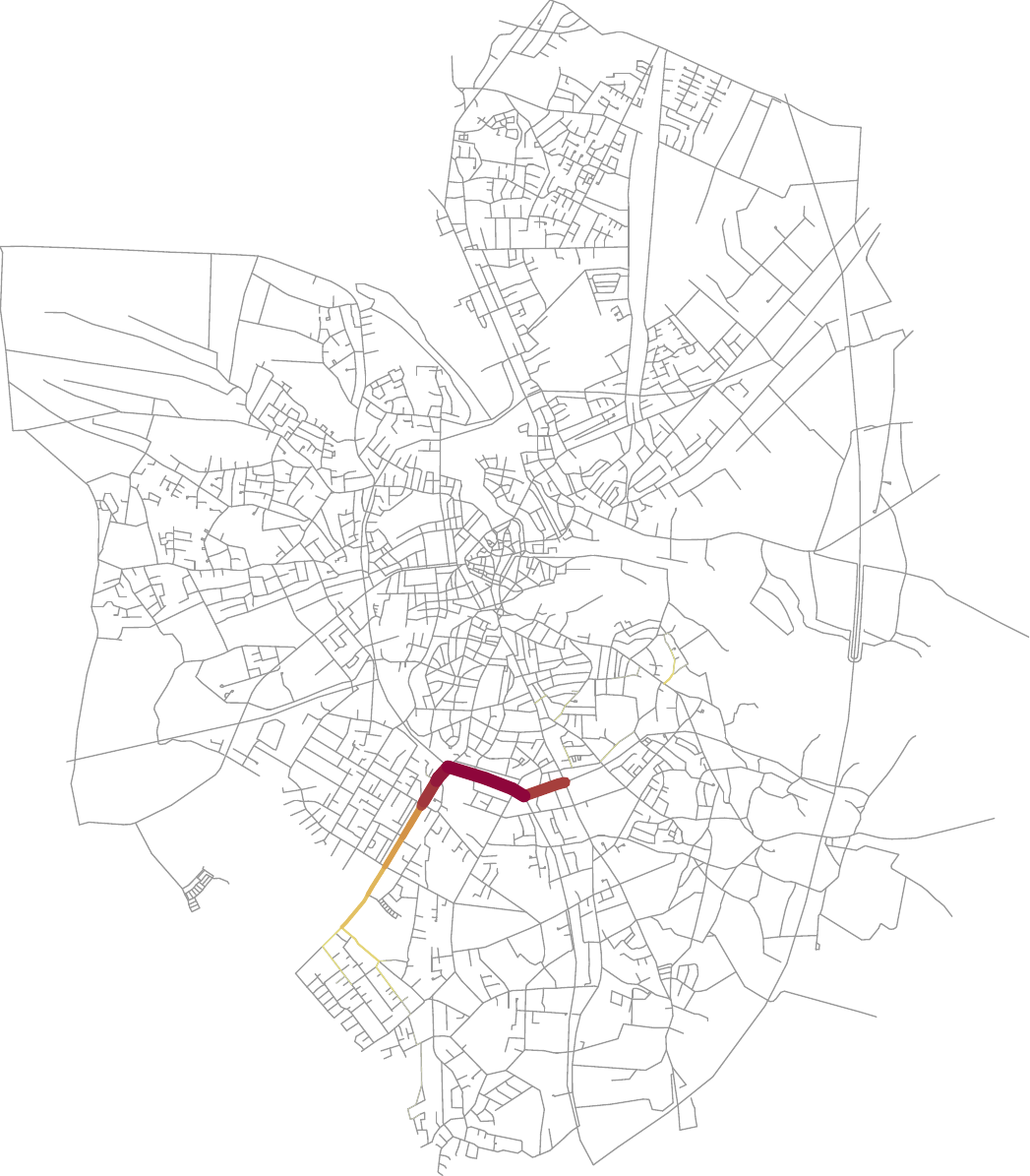}
}
\subfigure[]{
		\includegraphics[scale=0.35]{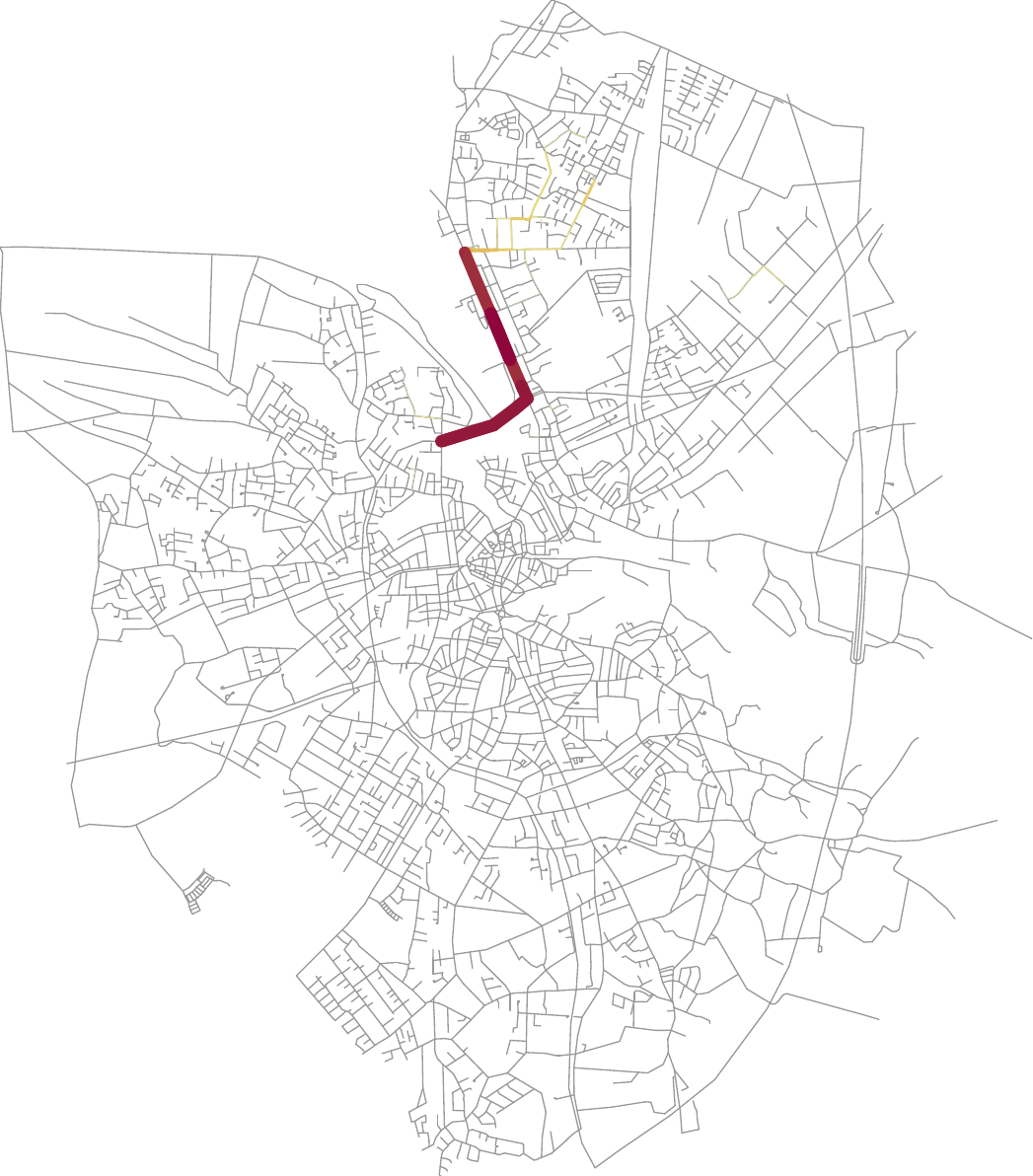}
}
\subfigure[]{
		\includegraphics[scale=0.35]{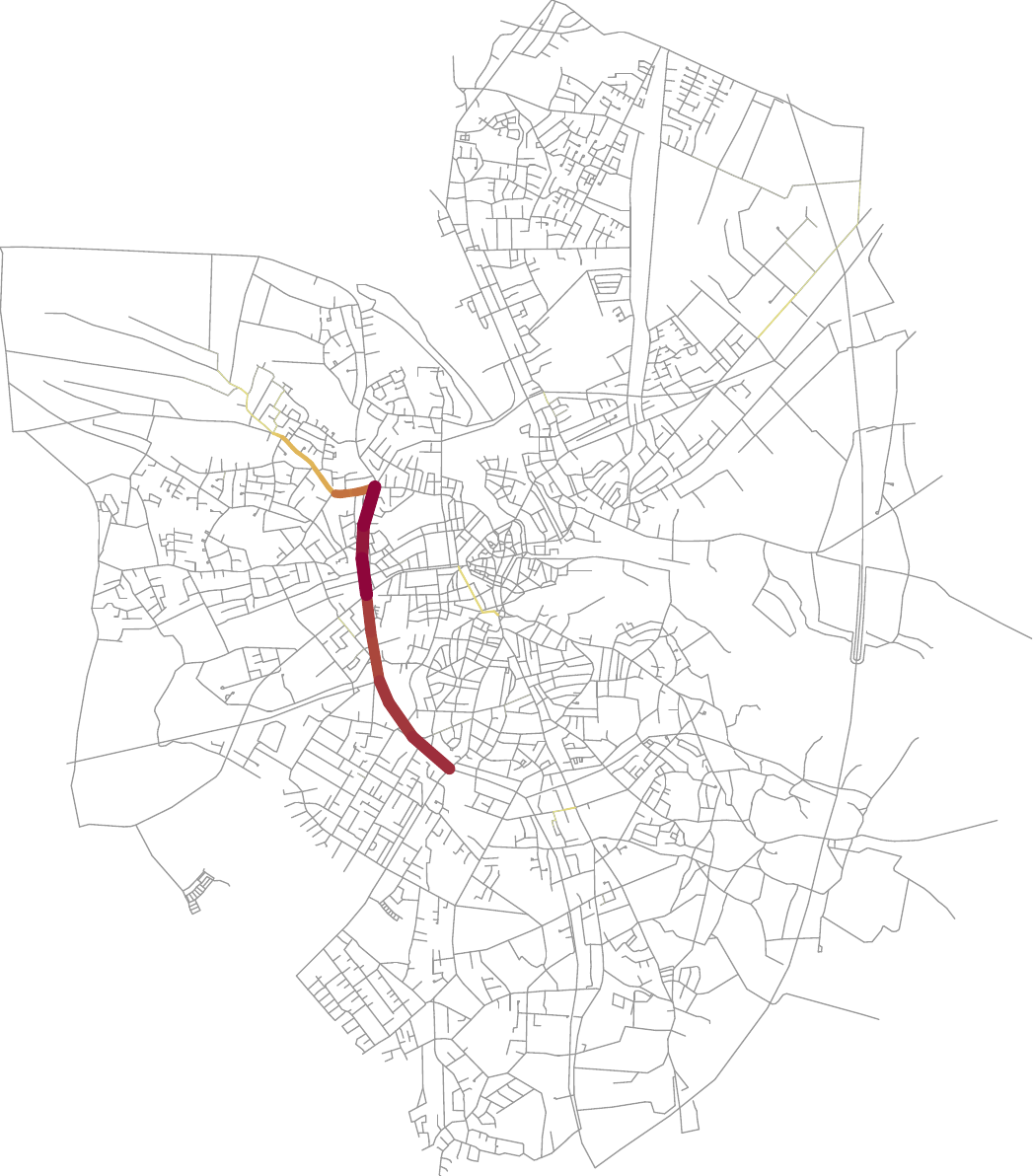}
}
\caption{Some clusters of segments discovered through clustering of the loose similarity graph. Clusters depicted in (a), (b) and (c) are top-level, very coarse clusters. Clusters (d) through (f) illustrate how clusters get refined when navigating down the hierarchy of clusters (in this case, the clusters are issued from the second level). The segments' coloration is relative to the number of trajectories traveling them, varying from pale yellow (for less traveled segments) to dark red (for segments that are traveled frequently).}
\label{fig:loose}
\end{figure}

Applying the loose approach results in a similarity graph containing 572903 edges. Clustering this graph yields a hierarchy of clusters that spans over seven levels, with only 16 discovered clusters at the top-most level and up to 1222 clusters in the bottom level. Clusters found at the highest levels are, generally, very coarse and are hard to interpret as they regroup a very large number of road segments that can be scattered across different regions of the road network (cf. clusters (a) through (c) in Fig.\ref{fig:loose}). Cluster quality rises as clusters split into more meaningful sub-clusters in the sub-sequent levels of hierarchy (clusters (d) through (f) in Fig.\ref{fig:loose}).

\begin{figure}[h]
\centering
\subfigure[]{
		\includegraphics[scale=0.35]{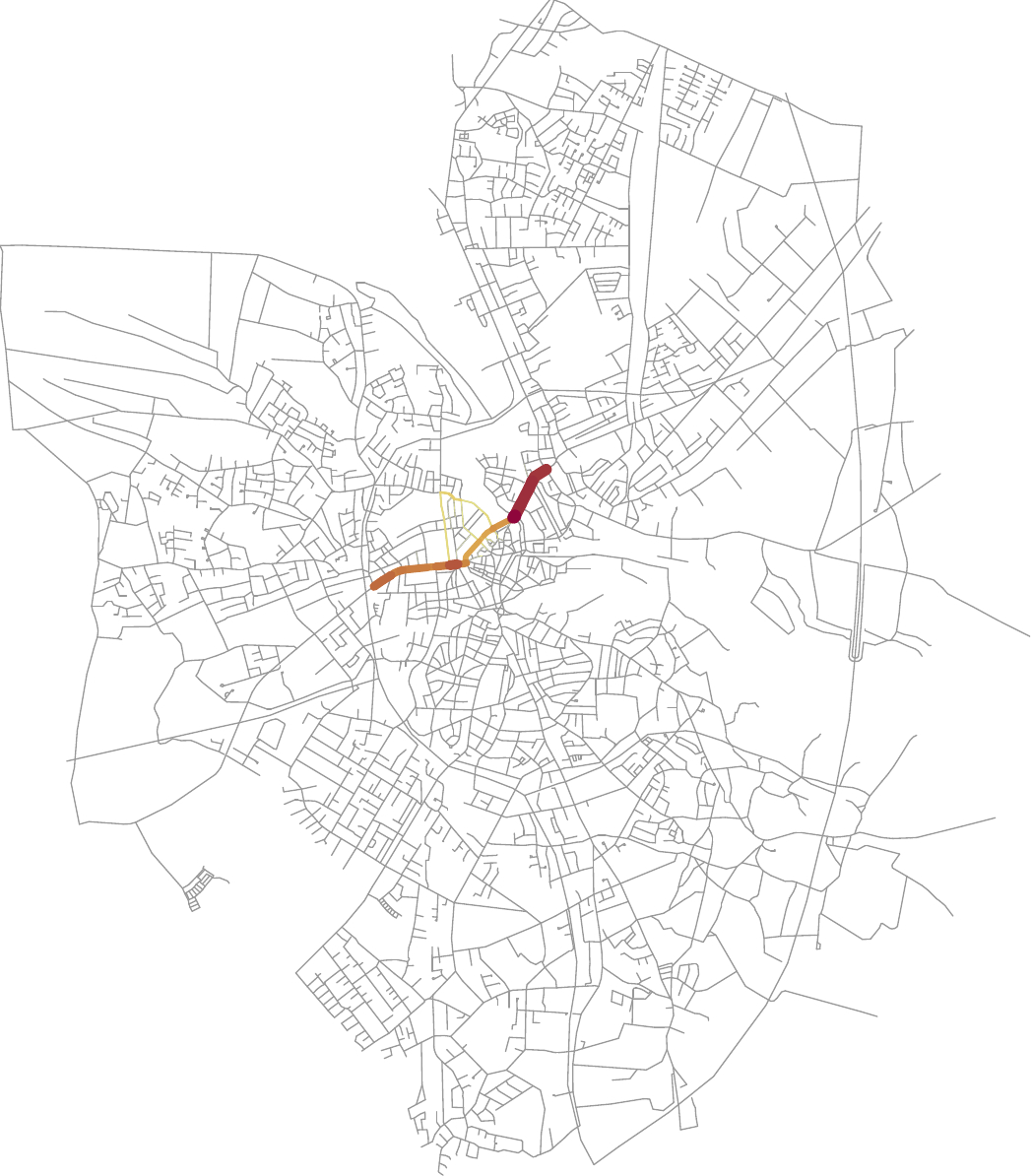}
}
\subfigure[]{
		\includegraphics[scale=0.35]{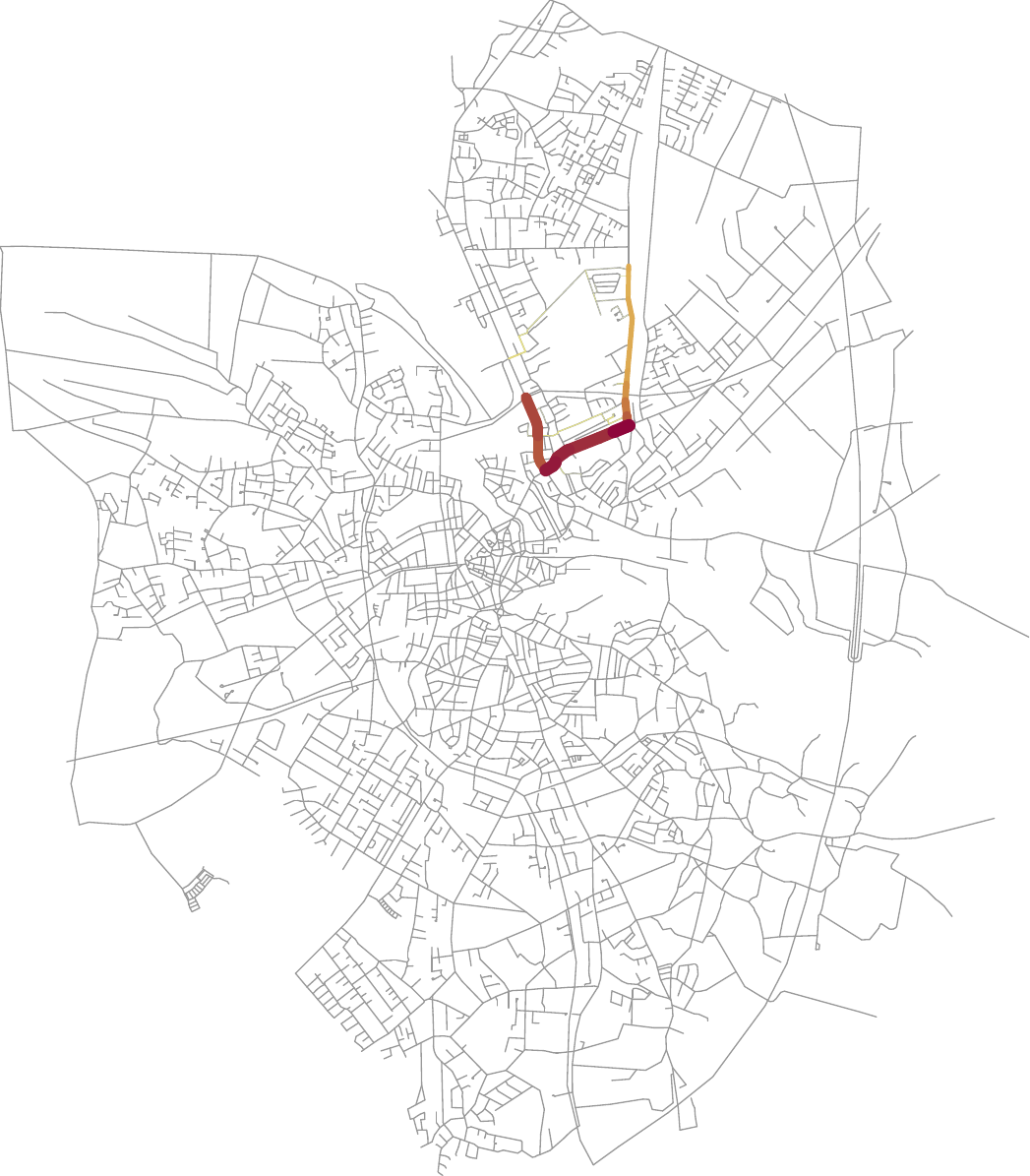}
}
\subfigure[]{
		\includegraphics[scale=0.35]{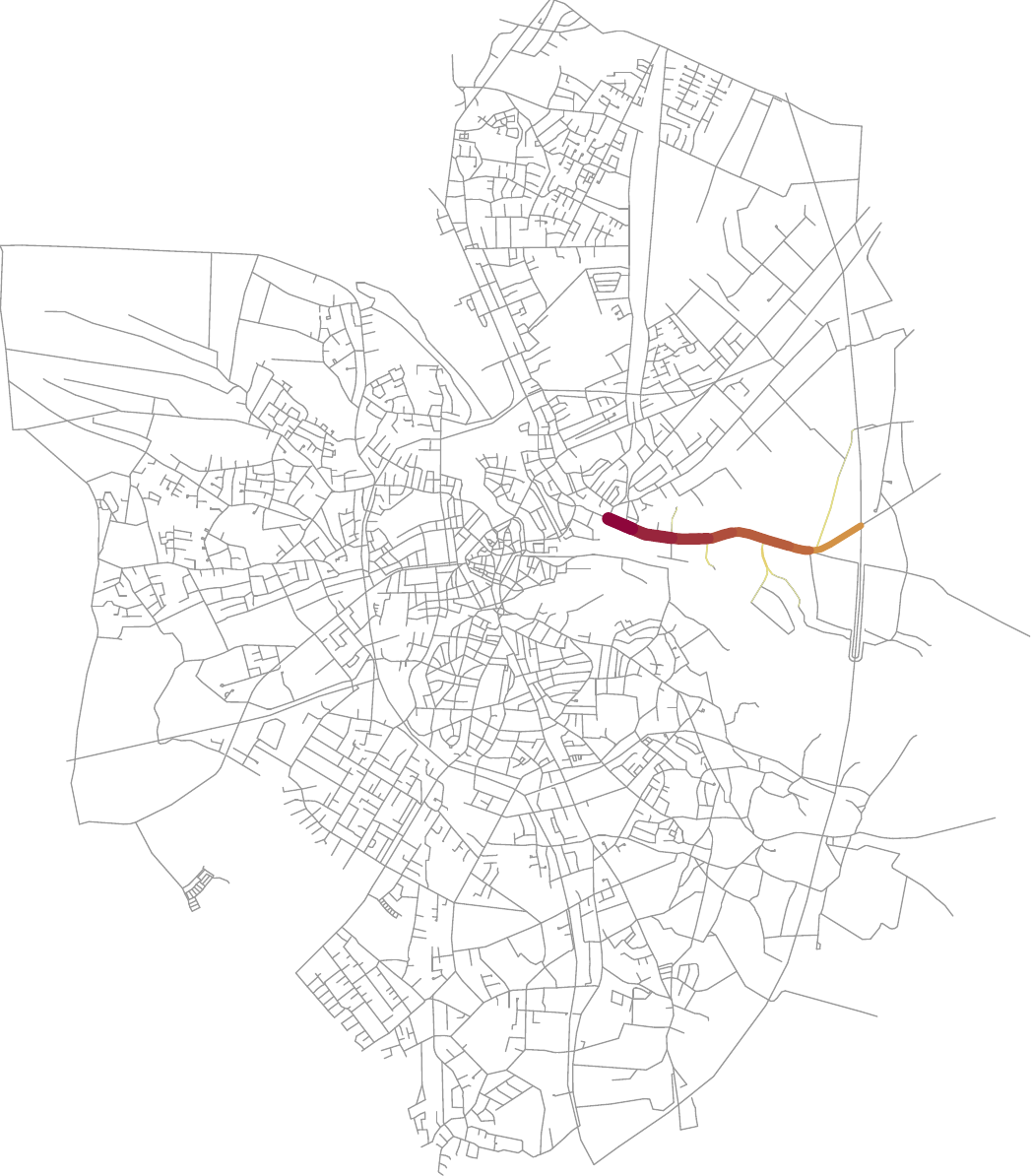}
}
\subfigure[]{
		\includegraphics[scale=0.35]{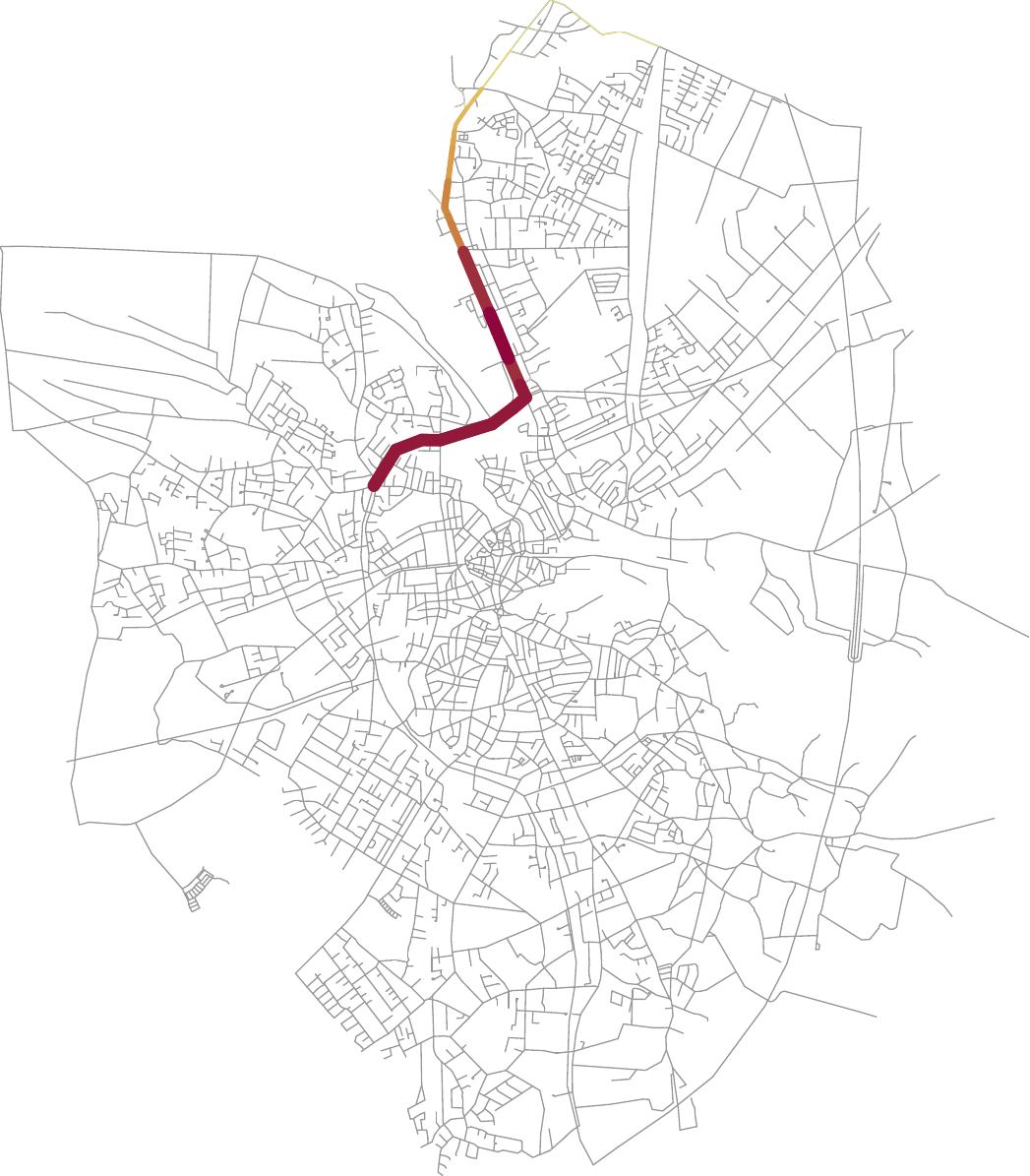}
}
\subfigure[]{
		\includegraphics[scale=0.35]{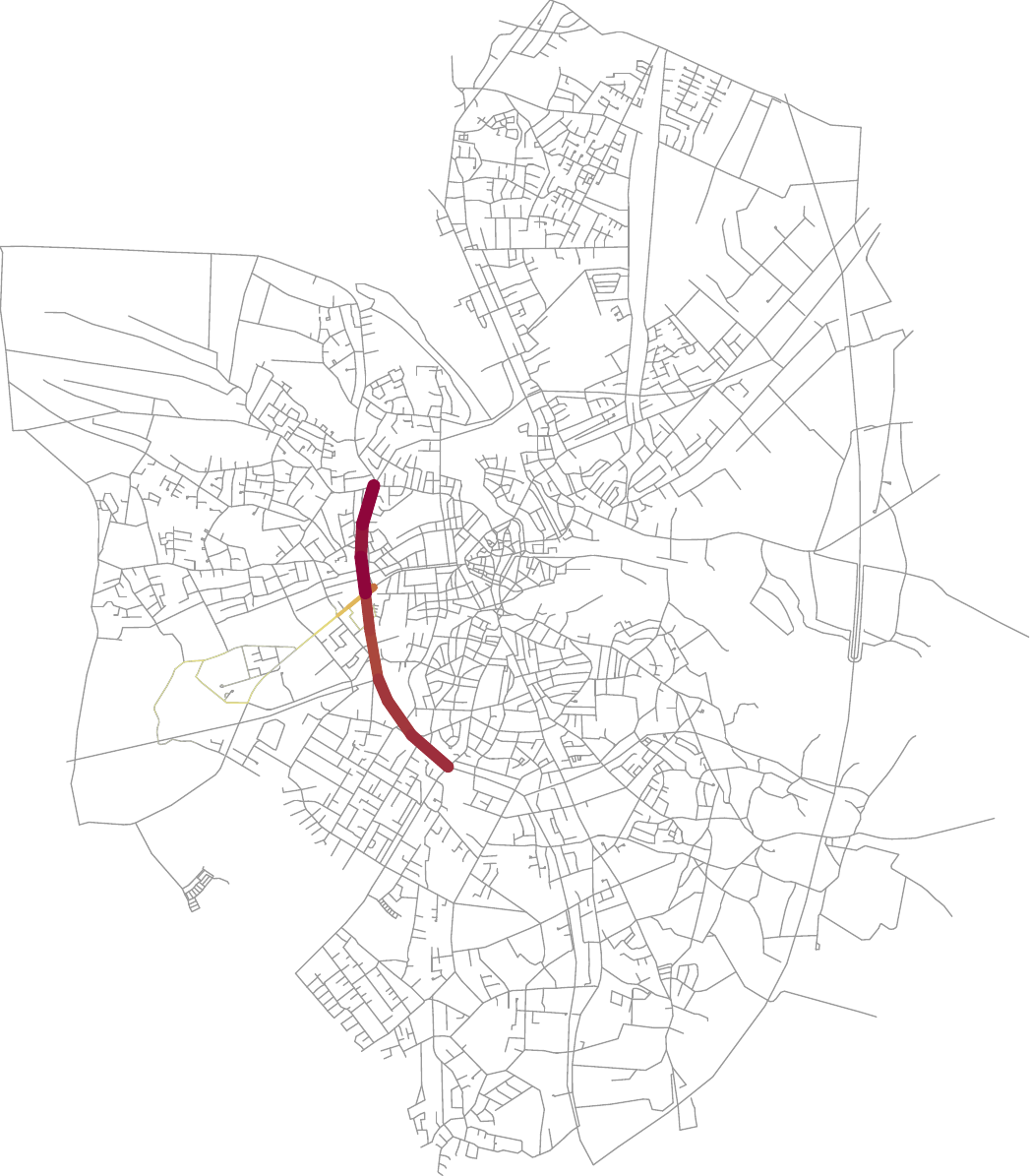}
}
\subfigure[]{
		\includegraphics[scale=0.35]{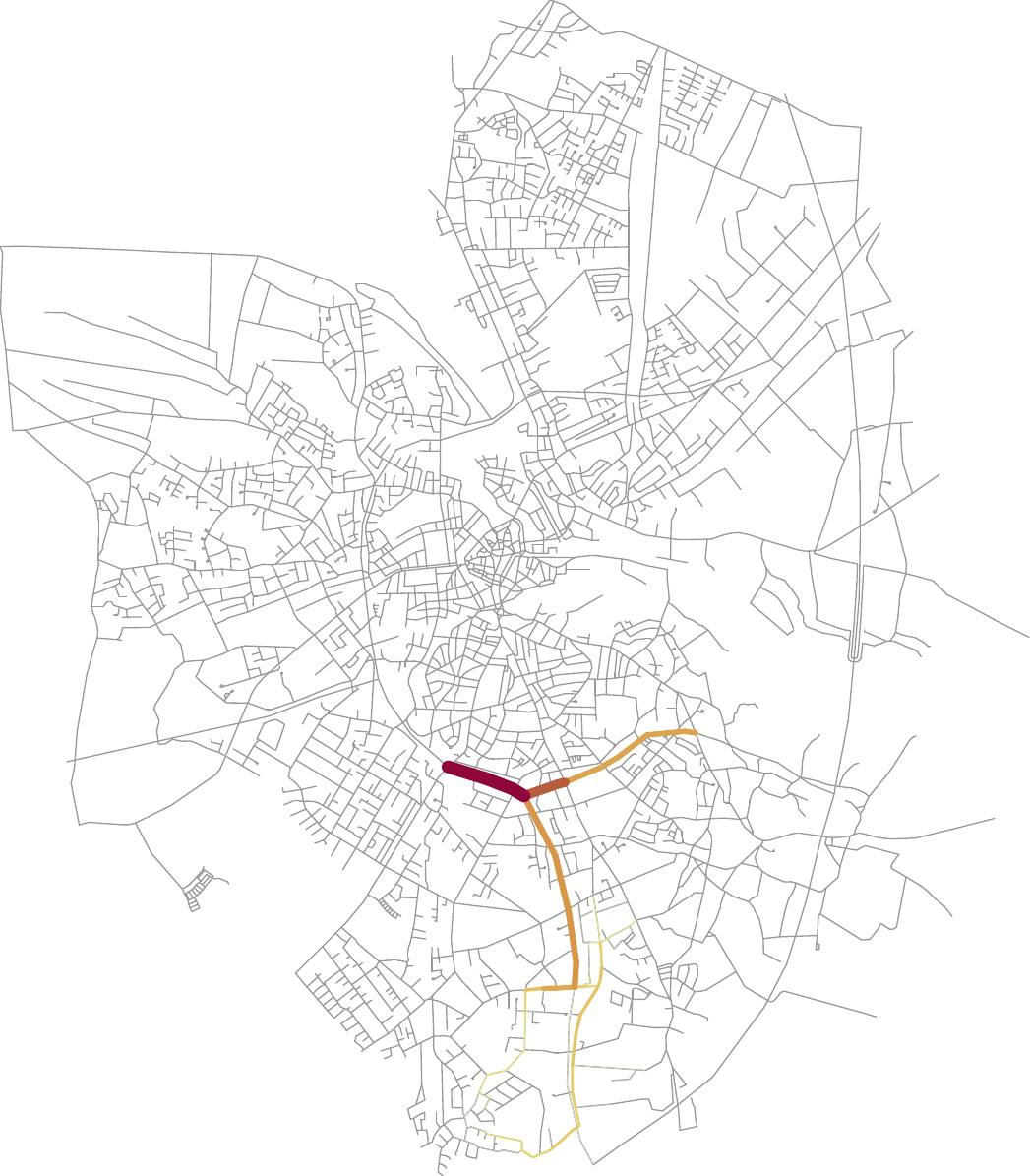}
}
\caption{Some of the clusters discovered through clustering of the strict similarity graph.}
\label{fig:strict}
\end{figure}

The strictly constructed graph, on the other hand, contains only 8674 similarity edges and is clustered by the community into a hierarchy containing only three levels. The top-most level is composed of 92 levels whereas the third level contains 216 clusters. Fig.\ref{fig:strict} showcases some of the top level clusters obtained by application of the strict segment clustering.

For comparison's sake, we tried clustering the road segments using classic hierarchical agglomerative clustering  (HAC) with average linkage. However, this approach resulted in very disproportionate clusters of poor quality (visualization of resulting clusters is omitted due to space limitation). The reason is that, during the first merging steps, the HAC can make some poor choices that are not reconsidered in the following steps. This limitation is overcome by the community detection algorithm used during our clustering step (cf. Section \ref{sec:clustering}) which can permute cluster participants if it can positively contribute to the modularity of the resulting partition.

Further experimentations are needed in order to compare clusters produced by the loose approach to those produced by the strict approach and objectively evaluate their quality.

\section{Related Work}
\label{sec:RelatedWork}

Prior work can be divided into two main areas: i. study of trajectory distances and similarity measures; and ii. design of trajectory clustering algorithms.

Distance measures proposed for free flow trajectories include Dynamic Time Warping (DTW) \cite{Berndt_1996}, LCSS (Longest Common Subsequence) \cite{Vlachos_2002}, ERP (Edit distance with Real Penalty) \cite{Chen_2004}, EDR (Edit Distance on Real sequence) \cite{Chen_2005} and the One-Way Distance (OWD) \cite{Lin_2005}. The above distances  consider trajectories as sequences of points on the euclidean space and ignore any network-related constraints that can restrict the movement of the studied objects. Therefore, they cannot be applied directly in our context. A constrained approach is presented in \cite{Hwang_2005}. However, it requires a priori definition of points of interest in the road network and can't be used for unsupervised learning. Other propositions can be found in \cite{Tiakas_2006} and \cite{Chang_2007}.

Approaches to trajectory clustering are adaptations of existing algorithms. Existing problem formulations and propositions include flock patterns \cite{Benkert_2006}, convoy patterns \cite{Jeung_2008a}, the TRACLUS partition-and-group framework \cite{Lee_2007} and the T-OPTICS and TF-OPTICS algorithms \cite{Nanni_2006}. The aforementioned algorithms use euclidean-based similarities and distances and can, therefore, be used only in the case of unconstrained trajectories. Furthermore, the majority of these approaches use density-based algorithms which suffer from two major drawbacks: i. their results are very sensitive to the parameter values; and ii. they assume that trajectories in the same cluster have a rather homogeneous density, which is rarely the case (as discussed in \cite{Roh_2010}).

In \cite{Kharrat_2008}, the authors describe an approach to discovering "dense paths" or sequences of frequently traveled segments in a road network. This approach resembles our segment-based clustering although they diverge on many key aspects. For instance, the approach in \cite{Kharrat_2008} produces flat clusters using a density-based approach (which requires fine tuning) whereas ours produces a hierarchy of nested clusters and does not require parametrization.

The most relevant work to our trajectory-based clustering is the one presented by Roh et Hwang in \cite{Roh_2010} where the distance between trajectories in the road network is measured using shortest path calculations. A baseline algorithm, using agglomerative hierarchical clustering, as well as a more efficient algorithm, called NNCluster, are presented for the purpose of regrouping the network constrained trajectories. Unlike our approach, the distance in \cite{Roh_2010} can be computationally prohibitive (especially in large road networks) besides from being direction-independent (i.e. it does not take into account the traveling direction, thus requiring an additional filtering step to separate inverted trajectories regrouped together).

\section{Conclusion}
\label{sec:Conclusion}

In this paper, we presented two approaches to cluster trajectory data under constraints of an underlying road network. The first approach focuses on clustering entire trajectories into communities of trajectories that exhibited similar behavior and movement patterns. The second approach deals with road segments and tries to discover groups of segments that are often visited concomitantly. Both approaches are based on a two-steps framework. First, they start by computing a similarity graph between the individuals to be clustered (i.e. trajectories or road segments). Then, the graph is clustered using a hierarchical community-detection algorithm.

This framework presents many advantages: i. it does not require parameters, contrary to the majority of existing approaches that are very sensitive to their threshold values; and ii. it also produces a hierarchy of nested clusters promoting exploration at various levels of granularity and detail in situations where a flat clustering approach would have produced a unique level containing a very large number of clusters. However, the framework is not flawless: the community detection algorithm used in the clustering step can be sensitive in presence of noise (i.e. trajectories that do not necessarily belong to any cluster or segments that are traveled rarely) which can degrade the quality of the discovered clusters.

For future work, we would like to conduct more thorough experiments with our segment-based approach and evaluate the quality of the produced clusters using internal and external quality indexes. We would also like to test our approaches on real vehicle trajectory datasets and study the impact of varying the community detection algorithm used in the clustering phase.

\bibliographystyle{splncs}
\bibliography{bibliography}

\end{document}